# Knowledge Graph Assisted Automatic Sports News Writing

Yang Cao[1], Xinyi Chen[1], Xin Zhang[2], Siying Li[2]
[1]School of Computer Science and Technology, Chongqing University of Posts and Telecommunications
[2]School of Information Science and Engineering, Chongqing Jiaotong University

**Abstract:** In this paper, we present a novel method for automatically generating sports news, which employs a unique algorithm that extracts pivotal moments from live text broadcasts and uses them to create an initial draft of the news. This draft is further refined by incorporating key details and background information from a specially designed sports knowledge graph. This graph contains 5,893 entities, which are classified into three distinct conceptual categories, interconnected through four relationship types, and characterized by 27 unique attributes. In addition, we create a multi-stage learning model by combining convolutional neural networks and a transformer encoder. This model expresses entity-task interactions using convolutional neural networks and enriches entity representations in the query set with the transformer encoder. It also includes a processor to compute matching scores for incomplete triples, addressing few-shot knowledge graph completion problem. The efficiency of this approach has been confirmed through both subjective and objective evaluations of 50 selected test cases, demonstrating its capability in revolutionizing the creation of sports news.

**0 Introduction**

At present, news automatic writing has gradually become one of the research hotspots, and many media and institutions at home and abroad have released their own writing robots by studying news automatic generation technology. In 2014, Quakebot [1], a robot of Los Angeles Times, automatically generated and published disaster news reports three minutes after the earthquake. Tencent Finance is the first media in China to use a news writing robot. The Dream Writer [2] developed by Tencent Finance automatically generates manuscripts in the first time. In November 2015, Xinhua News Agency began to try to use news automatic generation technology to write, and their "quick pen and small new" [3] can quickly generate a financial news. At the Rio Olympic Games in 2016, today's headlines realized the automatic generation of Olympic news by using the xiaomingbot [4] writing robot. It can be seen that news automatic writing technology has a very wide range of application scenarios, but it is still in the exploration stage because of the variety of news and the intersection of many different disciplines.

At present, there are three main development directions of news automatic generation technology, namely template, extraction and generation [5]. In recent years, there have been a series of representative studies on automatic writing in the field of sports news. Xu et al. [6] extracted important sentences from the live text broadcast of football matches by using text summarization algorithm to generate football news. Chen Yujing et al. [7] put forward a method to generate basketball match news according to the characteristics of basketball score changes in text live broadcast data. Zhu et al. [8] based on the conditional random field, combined with forward keywords and reverse keywords, extracted the key sentences from the live football text and automatically wrote the news report of football matches. In addition, for the automatic writing of football news, a series of methods [9-12] have emerged to automatically generate news scripts using convolutional neural network technology, which are essentially abstract or generative automatic summarization methods.

At present, the research on knowledge graph is very common, and there are many large-scale open-source knowledge graph libraries in China and abroad, such as Wiki data[13], Freebase[14], DBpedia[15], CNDBpedia[16], etc. In terms of application, the most typical application of knowledge graph is to assist search engine to improve the quality of search, such as Google Search of Google Inc. and Zhixin of Baidu. In addition, knowledge graphs can also be applied to Q&A systems to enhance the accuracy of the answer results. In addition to the applications of search engines and Q&A systems, recently there have been various research explorations of knowledge graphs in other fields, such as the field of intelligence [17], agro-tourism [18], and drug discovery [19], etc. Knowledge graphs related to journalism are also used in other fields, such as the field of information technology [20], the field of agriculture and tourism [21], and the field of drug discovery [22]. There are also many researches and applications of knowledge graph related to journalism, for example, Song Qing et al [20] analyzed the future application of knowledge graph technology in journalism; Reference [21] tried to use knowledge graph to detect the authenticity of news; and Reference [22] applied knowledge graph to improve the effect of text summarization. Reference [23-24] use knowledge graph for temporal logical reasoning.

To sum up, knowledge map can play a great role in many different fields, but there are few studies on automatic news writing. In addition, the existing large-scale open source knowledge map libraries are basically universal knowledge maps, which are often "breadth" but "depth" insufficient, and do not contain deeper information, so they cannot be directly applied to specific fields. Therefore, for the specific application of automatic writing of event news, it is necessary to build a specific knowledge map library in order to play a more effective role. The knowledge map has the way of searching knowledge, which is very suitable for automatically generating news by combining templates. In addition, knowledge map can provide more organized information for news audience. Structured entity knowledge makes news no longer superficial. If the knowledge behind news is presented in a visual form, it can enrich the automatically written news text, convey more information and provide in-depth reading of news.

In this paper, we constructed a knowledge graph of events reflecting news elements based on the characteristics of event news1, and proposed a method of automatic news writing that combines the fusion template of the knowledge graph, extracting and generating multiple modes of news. As shown in Figure 1, this paper first constructs a proprietary knowledge graph of events by defining the entities, attributes and relationships of news elements based on the crawled knowledge information in basketball-related fields. Next, we propose a KeyEventExtraction (KEE) algorithm to build a library of descriptive templates for highlighting game trends with key events to generate news drafts. Combined with the knowledge graph, this work can deeply explore the background information of players, teams, games, etc., and present users with rich and diversified visualized news to support in-depth reading.

**1 Construction of Knowledge Graph for Event News**

By constructing the knowledge map of events, this paper supplements the missing information such as game common sense and background, assists the generation of sports news, and makes the generated news articles of higher quality. The knowledge map of matches constructed in this paper belongs to a specific domain knowledge map, in which the entity class is defined as three important concepts in basketball matches, namely "team", "player" and "competition". For each entity, the basic information needed to assist the news generation of events is defined, such as "team" entity, and the attributes such as "name", "alias" and "partition" are defined. On the basis of attribute definition, this paper defines the relationship between entities that highlight the characteristics of news, including four relationships: current effectiveness, historical confrontation and participation in competitions. According to the relevant data crawled from Tigers Basketball Website 1 and Baidu Encyclopedia, we have constructed a knowledge map that can reflect the news characteristics of events. The statistical data are shown in Table 1. The total number of nodes in the knowledge map of events is 5,893, including 620 active players and 3,407 retired players. The number of team nodes is 30; From 1985 to 2019, there were 43,510 games, and the number of game nodes needed was 1,836.

1.1 Information Extraction

After completing the knowledge ontology modeling, in-depth knowledge extraction and management can be carried out on multi-source heterogeneous knowledge results, experience awareness and structured data, which mainly includes named entity recognition, relationship extraction, attribute classification and attribute value extraction.

Named entity recognition is the basis of information extraction, and its task is to find named entities from text,

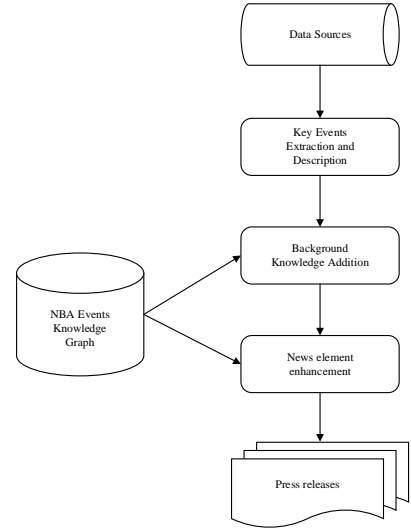

Fig. 1 Our Framework for Automatic Sports News Writing

which is unstructured data, and mark their types. In this paper, we use the natural language processing technology based on the combination of manual correction for named entity recognition.

Relationship extraction, on the other hand, requires extracting semantic relationships between two or more entities from text. Semantic relations are usually used to connect two entities and express the main meaning of the text together with the entities. Commonly, the results of relationship extraction can be represented by the subject-predicate-object (SPO for short, i.e., Subject, Predication, Object) ternary structure.

Attributes generally refer to the properties of entities or components that make up entities, and need to be extracted and categorized according to the composition of knowledge and the purpose of the task. Since these attributes are often evaluated by users in the text, they can be called opinion target extraction. Usually, attributes with the same semantics after extraction should be normalized.

Table 1 Statistics of knowledge graph results

| Statistical data | Numerical value |
| --- | --- |
| Number of entity classes | 3 |
| Number of Relationships | 4 |
| Number of attributes | 27 |
| Number of player nodes | 4 027 |
| Number of active player nodes | 620 |
| Number of retired player nodes | 3 407 |
| Number of teams | 30 |
| Number of games played | 43 510 |
| Number of match nodes | 1 836 |
| Number of summary points | 5 893 |

## 1.2 Knowledge Fusion

There are many data sources of knowledge map, and their naming specifications, entities and attributes have certain differences in their meanings. For this reason, in the process of constructing knowledge management system, in order to reduce the ambiguity of knowledge map, we should realize the integration and deep fusion of the unstructured knowledge results and structured data. Knowledge fusion includes the fusion of conceptual layer and entity layer, the fusion of conceptual layer is mainly based on knowledge expansion, and the fusion of entity layer adopts entity link technology. Firstly, based on the system knowledge formed by knowledge system classification, candidate entities are selected from various data sources through search engine; subsequently, supervised learning method is applied to train candidate entity ranking model by manually annotating the training set, and sorting candidate entities; after that, unsupervised learning method is applied to train the model based on unlabeled corpus, and amend and expand candidate entities; finally, entity similarity algorithm is used to complete the entity ranking model. After that, unsupervised learning method is applied to train the model based on the unlabeled corpus and correct and expand the candidate entities; finally, the entity layer fusion is completed by entity similarity algorithm.

## 2 Small Sample Knowledge Graph Completion Model Based on Meta-Learning

### 2.1 General Framework

Meta-TKGC consists of a relational meta-learner, a Transfomer encoder [25] and the matching processor. Fig. 2 shows the overall architecture of the model, and Fig. 3 shows the implementation of the relational metalearner. The model is based on CNN The relational meta-learner obtains task relation embeddings based on the reference entity pairs; the initialized task relation representations, query set and negative examples of the query set are spliced into a ternary group, combined with their positional information, respectively, and input into the Transfomer encoder to obtain the embeddings of the query set and its negative examples; task relation embeddings are relocated to the query part, and the head and tail scores obtained from the first two parts are computed in a matching processor to complete the small-sample knowledge graph complementation task. In the matching processor, the similarity scores between the entity and relationship feature representations obtained from the first two parts are computed to complete the small-sample knowledge graph completion task.

### 2.2 Relation meta-learner

Relationmeta learner is a two-layer feed-forward neural network, which consists of a convolutional layer and a fully connected layer to model the interaction between entities and relations. Firstly, the head and tail entities of the support set are spliced and input into the 2D convolutional layer, after activation and maximum pooling, it is reconstructed into a third-order tensor that integrates the interaction between entities and relationships in the reference set.

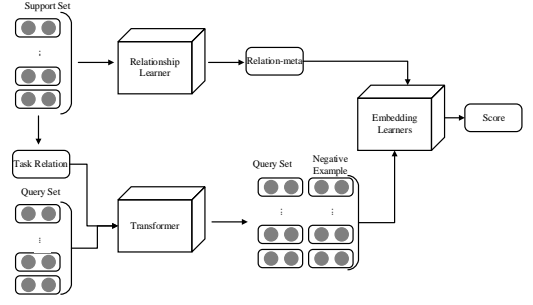

Fig. 2 Meta-TKGC Framewok

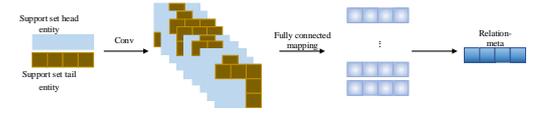

Fig. 3 Relation-meta learner

$$R_m^i = \sigma([S_h, S_t] * \omega) \quad (1)$$

where $S_h$ and $S_t$ denote the embedding of the head and tail entities of the support set, $\omega$ is the filter of the convolutional layer, the $\sigma$ denotes the ReLU activation function. The input is then fed into a linear layer for batch normalization:

$$R_s^i = W \cdot R_m^i + b, \quad (2)$$

where $W$ and $b$ denotes the learnable parameters. Secondly, for the parameter consisting of $K$ averaging the obtained relational representations over the entities in the reference set yields a task-relational embedding $R_s$ that can be migrated to the query set:

$$R_s = \sum_{i=1}^{K} R_s^i \quad (3)$$

### 2.3 Transfomer Encoder

Since the existing embedding models always need enough training data to model the task relations, but there is less relevant information to base on in practical applications, it is difficult to obtain effective representations of small-sample relations. Inspired by the translation distance model, based on the translation assumption, the embedding model is given by $h + r = t$ and we get $r = t - h$, thus computing the original embedding of the task relation from the head and tail entity embeddings of the support set $R_0$ for

$$R_0 = S_t - S_h \quad (4)$$

where $S_h$ and $S_t$ denote the embeddings of the head and tail entities of the support set, respectively. For the head and tail entities of the query set and its negative examples, the corresponding pre-trained embeddings and the ternary positional embeddings are summed to obtain the positional information fusion embeddings of the entities and relations:

$$h_i = h_0 + h_{\text{pos}} \quad (5)$$
$$t_i = t_0 + t_{\text{pos}} \quad (6)$$

where $h_0, t_0$ denote the original embeddings of the head and tail entities, respectively. $h_{\text{pos}}, t_{\text{pos}}$ represent the positional embeddings of the head and tail entities respectively. After obtaining the entity embedding, it is spliced with the relational meta-embedding obtained from the previous module to form a ternary embedding, thus realizing the effective combination of the reference set information and the query set information, and then input it into the L-layer transformer.

$$E_i^1 = \text{Transformer}(h_i, R_0, t_i), \quad (7)$$
$$E_i^l = \text{Transformer}(E_i^{l-1}), \; l = 2,3,\cdots,L, \quad (8)$$

where $R_0$ is the original embedding of the relation, $E_i^l$ denotes that after the L-Layer Transformer processed feature representation of an entity or a relation. In order to obtain the embedding of entity pairs in incomplete triples, the final hidden state is used as an expectation representation for query entity pairs and negative case entity pairs. This feature representation encodes the semantic roles of each entity, which helps to identify the fine-grained meanings of the task relations associated with different entity pairs.

2.4 Matching Processor

According to the score function $h + r = t$ in the pure translation model TransE, calculate the score function to measure the accuracy of triples:

$$\varphi(H_j, R_s, T_j) = H_j + R_s - T_j \quad (9)$$

where: the applied relational element $R_s$ is the result of the first part of the model, that is, the relational element embedding obtained by the support set; the entity embedding $H_j, T_j$ is the result of the fact that $L$ computed by the layer Transformer encoder. The loss function of the model is defined as

$$L = \sum_r \sum_{(H_q,T_q)\in Q_r} \sum_{(H_n,T_n)\in Q_r^-} [\varphi(H_q, R_s, T_q) - \varphi(H_n, R_s, T_n) + \gamma]_+, \quad (10)$$

where: $[x]_+ = \max\{0, x\}$, $\gamma$ to distinguish the interval between positive and negative examples, the $Q_r, Q_r^-$ denote the query set and the set of negative examples of the query set, respectively. During the training process, the purpose of distinguishing between positive and negative examples is achieved by increasing the interval between the scores of positive and negative samples of the query set, where the negative examples of the query set are constructed by randomly replacing the tail entities in the entity pairs.

**3 Automatic News Writing**

In order to generate relevant news, this paper proposes a KEE algorithm, which extracts key events from the live broadcast data of the games, then formulates a news template according to the characteristics of news, and generates a first draft of the news for the key events extracted by the KEE algorithm. On the basis of the first draft written by the KEE algorithm, this paper combines the constructed knowledge graph of games to supplement and improve the content of the first draft, so as to obtain higher-quality news results.

3.1 Extraction of Key Events

In the live text broadcast, each piece of data records an event that happened on the court, but these individual events cannot all appear in the generated news, so it is necessary to summarize and select the key events to generate news content. To obtain key events, the text is segmented to summarize the key events that occurred at different times, and then the news elements are extracted.

Because the key events in the competition are often accompanied by the change of the score, this paper segments the data by analyzing the change trend of the score. For each piece of information $L$ in the text live broadcast data, it can be expressed in quintuple (quarter, time, team, event, score). Among them, quarter refers to the current number of quarters, time is the remaining time in this section, team is the team where the event happened, event is the event that happened, and score is the score after the event. Remember that the difference between the scores of the home team $a$ and the visiting team $b$ at a certain moment $t$ in the basketball match is $\text{dif}(t)$, which can be expressed by the score at the moment $t$ in the live broadcast data as shown in Formula (11).

$$\text{dif}(t) = \text{score}_{a,t} - \text{score}_{b,t} \quad (11)$$

Among them, at time $t \in \{t_0, t_1, \cdots, t_n\}$, the score difference sequence set $\{\text{dif}(t_0), \text{dif}(t_1), \cdots, \text{dif}(t_n)\}$ at all times can be obtained. Here, it is described from the perspective of the home team. A positive score difference means that the home team is ahead, and on the contrary, a negative score difference means that the home team is behind. In order to obtain the development process and trend of news audience's interest in an event, it is necessary to further analyze the frequency and amplitude of the difference fluctuation from the obtained result.

In this paper, a segmentation method based on key time points is constructed to weigh the frequency and magnitude of spread fluctuations. Since the moment when the spread reaches the maximum and the moment when the spread reaches the minimum in an game are both a turning point of the game, we define these two moments as the key time points of the game, which are represented by $key\ time_1$ and $key\ time_2$ respectively, as shown in Eq. (12) and (13).

$$\text{key\_time}_1 = \arg\max_t \text{dif}(t) \quad (12)$$
$$\text{key\_time}_2 = \arg\max_t \text{dif}(t) \quad (13)$$

Next, it is necessary to judge the interval according to two key time points: if the key time point appears in the first sixth time of this competition, the start time $t_0$ of this competition is taken as the key time point; If it appears in the last sixth time of this competition, the end time $t_n$ of this competition will be regarded as the key time point. Therefore, $key\ time_1$ and $key\_time_2$ can be expressed as shown in Formula (14).

$$\text{key\_time}_i = \begin{cases} t_0, & \text{key\_time}_i \geqslant t\dfrac{n}{6} \\ \text{key\_time}_i, & t\dfrac{5n}{6} < \text{key\_time}_i < t\dfrac{n}{6} \\ t_n, & \text{key\_time}_i < t\dfrac{5n}{6} \end{cases} i=1,2$$

(14)

The following will be divided into $dif(key\_time_1) - dif(key\_time_2) > 8$ and $dif(key\_time_1) - dif(key\_time_2) \leq 8$ to judge the overall trend of this competition. When the score difference is greater than 8, the score difference fluctuates greatly, which can be regarded as a big difference before and after the key time point. At this time, this game is suitable to be cut into three segments according to the time point, which can be summarized as the following situations according to the situation on the field:

(1) $key\_time_1 > key\_time_2$, $dif(key\_time_1) > 0, dif(key\_time_2) < 0$: first "expand the advantages" and then "be overtaken" and then "narrow the disadvantages";

(2) $key\_time_1 < key\_time_2$, $dif(key\_time_1) > 0, dif(key\_time_2) < 0$: first "expand the disadvantage" and then "overtake" and then "narrow the advantage";

(3) $key\_time_1 > key\_time_2, dif(key_{time_1}) < 0$: first "narrow the disadvantage" and then "expand the disadvantage" and then "narrow the disadvantage";

(4) $key\_time_1 < key\_time_2$, $dif(key_{time_1}) < 0$: first "expand the disadvantage" and then "narrow the disadvantage" and then "expand the disadvantage";

(5) $key\_time_1 > key\_time_2$, $dif(key\_time_2) > 0$: first "expand the advantage" and then "narrow the advantage" and then "expand the advantage";

(6) $key\_time_1 < key\_time_2$, $dif(key\_time_2) > 0$: First, narrow the advantage, then expand the advantage and then narrow the advantage.

Table 2: Segmentation and Trend Summary (Partial)

| Number of Sections | Segment Trend Summary | Segment Trend Description |
|---|---|---|
| 1 Segment | Steady lead | No further expansion of the home team's advantage and no sign of narrowing it |
|  | Stalemate | Both teams lead alternately |
| 2 segments | Overtaken - Reducing the deficit | The home team starts the game with a lead, then is tied and overtaken, and finally closes the gap. |
|  | Rebounded - Reduced Advantage | The home team starts the game with a deficit, then ties the game and closes the gap. |
| 3 segments | Increase Advantage - Be Overrun Reduce Disadvantage | The home team starts the game with a deficit and reaches the maximum difference, then is tied and overtaken, and finally closes the deficit. |
|  | Increase disadvantage - overtake decrease advantage | The home team falls behind by a minimum (negative) margin, then is tied and overtaken to take the lead, and finally the lead is reduced |

When the score difference is less than 8, the fluctuation range of the score difference of the competition in this section is small, and the situation of the stadium is biased towards a seesaw state. At this time, the competition in this section can be treated as a whole section, which can be divided into the following three situations:

(1) $dif(key\_time_2) > 10$: the minimum score difference is greater than 10 and the score difference does not change much, and the situation of the stadium is summarized as the home team's "steady lead";

(2) $dif(key_{time_1}) < -10$: The minimum score difference is less than -10 and the score difference does not change much. The situation of the stadium is summarized as that the home team is "steady and backward";

(3) $dif(key\_time_1) > 0, dif(key\_time_2) < 0$: the score difference changes around 0, and the situation of the stadium is summarized as "stalemate" between the two teams.

In addition to the above general situation, there are a few special circumstances, which are more suitable for dividing an game into two segments. Based on the above situations, we divide a competition into some situations as shown in Table 2, which correspond to the score trend of the home team and the visiting team respectively, which is helpful for the subsequent news generation.

According to the segmentation results of each match, the key events concerned by Chinese news reports can be extracted from all basketball match events in that match, such as event $t_t, t \in \{t_0, t_1, \cdots, t_n\}$. We first classify the basic events in the text live broadcast, and then convert the events to extract them, which are summarized as follows:

(1) Score categories: Score categories include "two-pointer", "three-pointer" and "free throw hit" in the live text broadcast.

(2) Strike iron categories: Strike iron categories include "two points miss" and "three points miss".

(3) Rebound category: Rebound category includes "offensive rebound" and "defensive rebound".

(4) Foul categories: Foul categories include shooting foul, personal foul, offensive foul, malicious foul and technical foul.

(5) Error category: The error category includes "passing error", "out of bounds" and "losing the ball".

(6) lineup adjustment.

(7) pause.

Next, the relevant information is stored in the above-mentioned corresponding categories. According to the event description features (Table 3), all kinds of event $_t$ are converted from all events, and the key event $E^*$ is extracted, which is defined as five-tuple $E^*$ = (*player, team, action, result, time*), where *player* and *team* refer to the actor and his team; *action* is a concrete action, mostly a basic event; *result* is the score after the action; *time* is the time when the action takes place.

The key event extraction algorithm (KEE algorithm) proposed in this paper is as follows:

3.2 Description of key events

After extracting the key news events mentioned above, it is necessary to describe the key events to become part of a news report. In order to meet the description needs of key events, we have constructed corresponding templates. Table

Table 3 Key Events of Games (Partial)

| Event Name | Performer | Event Description |
|---|---|---|
| Highest Score | Player | The player who has helped his team score the most points in a given period of time. |
| Consecutive Points | Player | A player who has scored consecutive points to help his team over a certain period of time. |
| Significant Score 1 | Player | At some point, a player scores to help his team get back in the game. |
| Important Score 4 | Players | Scoring when the team has not scored for a long time |
| Key Rebounds | Player | Rebounds grabbed before key points |
| Most Turnovers | Player | Most turnovers in a period of extended disadvantage |
| Most iron shots | Players | Most shots thrown during a stretch of extended possession |
| Offensive Highlights | Team | Number of points scored by the team in the period before the team reaches the maximum lead |

Table 4: Templates for describing key events (part)

| Event Name | Template |
|---|---|
| Highest Score | [#Player] started to explode, and he cut down [#Scores] by himself, helping [#Team] to overtake the score. |
| Continuous Scoring | [#Player] has a hot streak of [#Scores]. |
| Important Score 3 | At the critical moment, [#Player] scored [#Points] to save [#Team]. |
| Key Rebound | The [#Player] grabs the key rebound and saves the game. |
| Most Iron Shots | [ # Player] is in a slump and misses a series of shots, putting [ # Team] at a disadvantage. |
| Key Timeout | [ # Team] calls a key timeout, which is the turning point of the game. |

4 gives some key event description templates.

In addition to the key events defined above that need to be covered in the story, a description of the overall trend of a game can also be part of the story, as shown in the Game Trend Description template in Table 5.

Generating news generally requires a combination of two types of templates in order for the final news to be effective. To do this, randomly select key events to insert into the corresponding segments of the trend template, and generate the following example:

"The Pelicans started the game with a slight advantage, leading the 76ers by about two points. However, the 76ers not only tied the game, but took the lead, leading by as many as 16 points, with Joel Embiid leading the way with 13 points. By the end of the first quarter, the 76ers led the Pelicans by 15 points."

In the knowledge map constructed in this paper, each team has at least one leader player and several star players.

Table 5: Template for describing trends (partial)

| Trend | Template |
|---|---|
| Sticking | [ # Team A] and [ # Team B] are locked in a tight battle, with the score moving up and down |
| Steady lead | [ #Team A] is firmly in the driver's seat, leading the [#score] by about one point. |
| Increased lead - overtaken - reduced deficit. | [ #Team A] came up and took the lead, leading [ #Team B] by [#Score], and then [ #Team B] rallied and not only evened out the difference, but also took the lead over [ #Team A] [#Score]. Toward the end of the period, [#Team A] was still trying to close the gap, and the score was getting closer. |
| Overtaken - Reducing deficit | With the lead from the previous quarter, [#Team A] failed to further increase its advantage and [#Team B] seized the opportunity to overtake the lead. At the end of the quarter, [#Team A] slowly got closer. |

It will make the news more vivid and topical to describe the game with the focus on such players and track and summarize the data of these people. The leader players and star players of the Los Angeles Lakers found in the knowledge map, and all their scoring events, rebounding events and assists events are obtained to generate summary data, and the following news report on player performance can be obtained:

" On the Lakers side, leader LeBron James had 17 points, 13 rebounds and 5 assists, star player Kyle Kuzma had 19 points and 6 rebounds, and Brandon Ingram had 14 points, 2 rebounds and 2 assists, which helped the team immensely. "

**4 Experimental results and analysis**

4.1 Data Preprocessing

In order to generate news about related events, it is necessary to write the live text data of related games crawled by crawler, and then preprocess the crawled data. The data used to generate news in this paper is concise text live broadcast data 1, which covers the development process of the whole game and is normative, although it does not contain the description of the details of the event. Table 6 is an example of a simple text live broadcast. In the concise basketball text live broadcast data, each piece of information can be represented by quintuple = (quarter, time, team, event, score). The types of events here are fixed, which generally include seven situations: pause, mistake, rebound, score, strike while the iron is hot, lineup adjustment and foul, and the events are independent, and there is no necessary connection between the events before and after.

4.2 Results and analysis of game news automatic writing

In this paper, we select the text broadcast data of games in Hupo Basketball website as the data source, and each game consists of at least 4 subsections. Firstly, we segment the data according to the live broadcast, and then automatically write the news according to the KEE algorithm combined with the template library. Based on the live streaming data of a game on HuffPost.com, we automatically generate a news release,

Table 6 Example of a Simple Text Broadcast

| Quarters | Time | Team | Event | Score |
| --- | --- | --- | --- | --- |
| First quarter | 11:44 | 76ers | Jimmy Butler, two points, miss. | 00 |
| First Quarter | 11:33 | Pelicans | Wesley Johnston with a rebound. | 00 |
| ... | ... | ... | ... | ... |
| First Section | 17.0″ | Pelicans. | Nikola Mirotic misses a three. | 2336 |
| First Quarter | 0.0 | Pelicans | Joel Embiid makes a two-point field goal | 2338 |

as shown in Fig. 5.

On the basis of the above results, this paper combines the knowledge graph of games to enhance the quality of the above press release, and on the basis of the above results, we can get the press release shown in Fig. 6, with the horizontal line highlighting the content of the press release generated by combining the knowledge graph. Comparing the two press releases, it can be seen that the quality of the press release combined with the knowledge graph is obviously higher, and the information supplementation effect brought by the knowledge graph is obvious.

In terms of comparison experiments, this paper uses the method of automatically generating soccer news using Convolutional Neural Network (CNN) as a comparison experiment. Since the above work does not disclose the dataset and source code of soccer text broadcasting, the evaluation adopts the BiLSTM+CNN method to conduct experiments on the basketball text broadcasting dataset that is disclosed in this paper, which is referred to as the CNN method in the following.

We randomly selected 50 games from the most recent season, and automatically generated CNN press releases, first press releases using the KEE algorithm, and final press releases combining the KEE algorithm and knowledge graph, and conducted subjective and automatic evaluations, respectively.

4.2.1 Subjective Evaluation

We invited 20 evaluators to participate in the subjective evaluation of news quality. Among these 20 evaluators, 4 are first-line sports journalists, 10 are basketball fans, and 6 are ordinary people who have only heard about events. Firstly, the overall quality of news is evaluated, and the evaluation criteria are divided into four levels: excellent, good, passing and poor. The evaluation results of different groups of people for abstract press releases are shown in Figure 4, and the evaluation results for the first draft of KEE news and the final draft of KEE news with knowledge map spectrum are shown in Figure 5 and Figure 6 respectively.

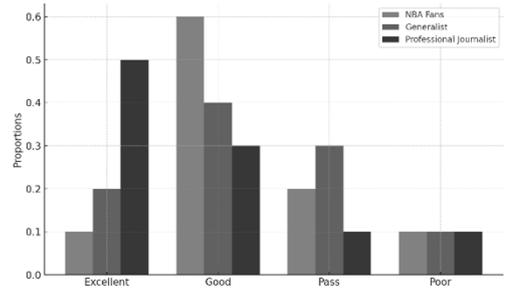

Fig. 4 CNN method press release overall evaluation results

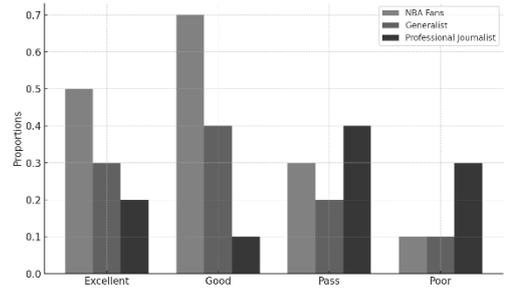

Fig. 5 Overall evaluation results of press releases by KEE method

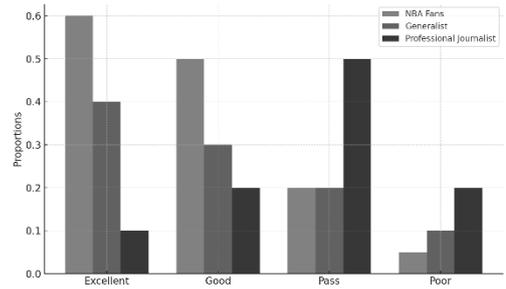

Fig. 6 Overall evaluation results of press releases with KEE method integrating knowledge graphs

The evaluation results show that CNN press releases have the lowest excellent and good rates, followed by KEE method, while the excellent and good rates have increased significantly after the integration of KEE method with knowledge mapping (for fans, the excellent rate increased from 17.80% to 41.60%; for professional journalists, the good rate increased from 39.50% to 52%). In conclusion, compared with the CNN method, the news generated by the fused knowledge graph in this paper is more satisfactory. Our methodology does not reach the level of professional journalism, but it does meet the needs of people, especially fans, who are direct readers of news in this field.

In order to provide a more comprehensive picture of the quality of the news and to compare the advantages of the knowledge map, we have set up five subjective evaluation indicators: Smoothness of expression, clarity of content, richness of information, naturalness of writing (whether or not there are obvious traces of machines), and reading experience.

Table 7  Subjective Evaluation Indicator Scores (Mean)

| | Fluency of expression | | | Clarity of content | | | Information richness | | | Naturalness of writing | | | Reading experience | | |
|---|---|---|---|---|---|---|---|---|---|---|---|---|---|---|---|
| | Abstract Format | KEE method | Combined with KG | Abstract | KEE Method | Combined with KG | Abstract | KEE Method | Combined with KG | Abstract | KEE Method | Combined with KG | Abstract | KEE Method | Combined with KG |
| Fans | 3.94 | 3.87 | 4.21 | 3.96 | 3.8 | 4.28 | 3.92 | 3.67 | 4.31 | 3.99 | 3.83 | 4.22 | 3.89 | 3.78 | 4.19 |
| General population | 3.84 | 3.84 | 4.00 | 3.82 | 3.75 | 3.96 | 3.85 | 3.47 | 4.28 | 3.72 | 3.61 | 3.78 | 3.55 | 3.47 | 3.79 |
| Professional journalists | 3.13 | 3.16 | 3.31 | 2.95 | 3.39 | 3.45 | 2.87 | 2.59 | 3.66 | 3.32 | 2.99 | 3.18 | 2.79 | 2.35 | 3.45 |

Each indicator is scored from 1 to 5, with 5 being the most consistent with the indicator and 1 being the least consistent with the indicator. The evaluators did not know which news was generated in which way, and scored the news according to their own subjective feelings.

Table 7 shows the average scores of the 20 evaluators (the results are rounded to two decimal places). It can be seen that the CNN method scores slightly higher than the KEE method in most cases; however, the KEE method combined with the Knowledge Graph improves the scores of each indicator, especially the "content clarity" and "information richness". However, the KEE method combined with knowledge mapping improved the scores of each indicator, especially in "content clarity" and "information richness", which is due to the fact that the application of knowledge mapping increased the background information and the focus of the news descriptions in the generated press releases.

4.2.2 Automatic Evaluation

In addition to the intuitive reading experience evaluation mentioned above, we also use ROUGE as the automatic evaluation method, using ROUGEN's F1 as the evaluation index, mainly examining the adequacy and necessity of the text generation results. The dataset used for the automatic evaluation comes from the interactive text broadcast data of 494 basketball games crawled from Hupo, with a total of 58,745 sentences, which are labeled by semi-automatic annotation methods. In the evaluation dataset, 43211 sentences from 340 games are used as the training set and 15534 sentences from 154 games are used as the test set.

In this paper, we compare the CNN press releases generated from the text broadcast data of 50 randomly selected games, the initial press releases using KEE algorithm, and the final press releases combining KEE algorithm and knowledge graph, and the press releases of the same games written by professional journalists on Hupo as the quality reference data. Table 8 shows the evaluation results of the three methods on the evaluation dataset under the same ROUGE environment. It can be seen that the experimental results of the KEE method proposed in this paper are slightly higher than those of the classical digest CNN method, which is more in line with the human-written press releases and not as mechanical as the digest method. The result of KEE+KG method is the best, which shows that the final press release generated by combining knowledge map can enrich the news content of events and greatly enhance its readability. The results of automatic evaluation are basically consistent with the results of subjective evaluation, which shows the effectiveness of the automatic writing method based on knowledge map.

Table 8 Comparison of the automatic evaluation of press releases with the CNN method.

| | CNN method | KEE method | KEE+KG method |
|---|---|---|---|
| Rouge1 | 0.264 | 0.392 | 0.563 |
| Rouge2 | 0.065 | 0.138 | 0.372 |
| RougeL | 0.182 | 0.197 | 0.447 |

4.3  News Presentation with Knowledge Graphs

In this paper, the automatically generated game news incorporating the Knowledge Graph is finally presented on a website page, and other background knowledge from the Knowledge Graph can be applied to enrich the presentation of the news text. A common news page usually contains the text of a news story, game pictures or game videos (Fig. 10). However, such a presentation is too monotonous and does not allow readers to access additional information beyond the news. Instead, we tagged some of the keywords in the news text and clicked on the tagged text to display additional background knowledge in the form of a "pop-up window". The background knowledge of a player or a team is presented in the form of a force-driven graph (Fig. 11), which obtains the attributes of the player or team entity from the knowledge graph library constructed in Section 1, so that the user can quickly understand the player or team.

In addition, news elements such as background events are presented in the form of labels for in-depth understanding. In this paper, the combination of the knowledge graph with the presentation of news allows readers to obtain more background knowledge in addition to the coverage of the game through interactive clicking, which greatly enhances the reading experience of the readers.

## 5 Conclusion

In this paper, we have proposed an automated method for writing game news by integrating knowledge graphs. Firstly, we segmented the games based on key time points, and then we obtained the trend descriptions of the games. Based on this, we define some key events and their corresponding description templates according to the characteristics of basketball games, which are used to generate the first draft of the news. In order to solve the problems of missing background information and unclear descriptions of event news elements in the first draft, the constructed event knowledge graph is used to assist in the generation of the final press release, thus improving the quality of the generated news. The press releases generated by this method are smooth in expression, clear in content, rich in information, high in falsehood and good in reading experience. In addition, the knowledge graph constructed in this paper also helps to visualize the depth of news reports.